\documentclass[runningheads]{llncs}
\usepackage{graphicx}
\usepackage{amsmath,bm,amsfonts,amssymb,mathrsfs}
\usepackage{color}

\usepackage[utf8]{inputenc}
\usepackage[T1]{fontenc}
\usepackage[bookmarks=false]{hyperref}
\usepackage{url}
\usepackage{booktabs}
\usepackage{nicefrac}
\usepackage{microtype}
\usepackage{algorithmic}
\usepackage{bm}
\usepackage{grffile}
\usepackage{xspace}
\usepackage{etoolbox}
\usepackage{array}
\usepackage{lmodern}
\usepackage{colortbl}

\usepackage{multirow}
\hypersetup{colorlinks=true,hidelinks,bookmarks=false}

\newcommand{\eg}{\textit{e.g.}\xspace}

\makeatletter
\let\MYcaption\@makecaption
\makeatother

\usepackage{subcaption}

\makeatletter
\let\@makecaption\MYcaption
\makeatother  

\usepackage{tikz,pgfplots,pgfmath}
\usetikzlibrary{plotmarks,shapes.arrows,calc,decorations.text}
\usetikzlibrary{decorations.pathreplacing}

\newlength{\figurewidth}
\newlength{\figureheight}

\pgfplotsset{every axis/.append style={
  grid style={line width=0.6pt,dotted,gray}}}

\newcolumntype{P}[1]{>{\centering\arraybackslash}p{#1}}

\newcommand{\toptitlebar}{
  \hrule height 4pt
  \vskip 0.25in
  \vskip -\parskip%
}
\newcommand{\bottomtitlebar}{
  \vskip 0.29in
  \vskip -\parskip
  \hrule height 1pt
  \vskip 0.09in%
}
\newcommand{\appendixtitle}[1]{{\phantomsection\hsize\textwidth\linewidth\hsize %
  \vskip 0.1in \toptitlebar \centering{\Large\bf #1\par}\bottomtitlebar%
  \addcontentsline{toc}{section}{#1}}}

\renewcommand{\orcidID}[1]{}

\begin{document}
\pagestyle{headings}
\mainmatter

\title{Unstructured Multi-View Depth Estimation Using Mask-Based Multiplane Representation}
\titlerunning{Multi-view depth estimation using mask-based multiplane representation}

\author{Yuxin Hou\orcidID{0000-0001-7266-9502} \and Arno Solin\orcidID{0000-0002-0958-7886} \and Juho Kannala\orcidID{0000-0001-5088-4041}}
\institute{Anonymous institute}
\institute{Department of Computer Science, Aalto University, Espoo, Finland \\
           \texttt{\{yuxin.hou, arno.solin, juho.kannala\}@aalto.fi}} 

\maketitle

\begin{abstract}
  This paper presents a novel method, MaskMVS, to solve depth estimation for unstructured multi-view image-pose pairs. In the plane-sweep procedure, the depth planes are sampled by histogram matching that ensures covering the depth range of interest. Unlike other plane-sweep methods, we do not rely on a cost metric to explicitly build the cost volume, but instead infer a multiplane mask representation which regularizes the learning. Compared to many previous approaches, we show that our method is lightweight and generalizes well without requiring excessive training. We outperform the current state-of-the-art and show results on the {\sc sun3d}, {\sc scenes11}, {\sc MVS}, and {\sc RGBD} test data sets. 
  
\end{abstract}

\section{Introduction}
\label{sec:intro}

Multi-view stereo (MVS) aims at reconstructing depth (or disparity) maps from a collection of overlapping images, which is a fundamental problem in computer vision. Any progress in the field will have a direct impact on applications like augmented reality and self-driving cars. Conventional methods often use hand-crafted features and compute similarity between patches. However, these approaches may suffer from limitations of the features, especially regarding poorly textured or reflective regions. As deep convolutional neural networks (CNNs) have shown great success in many vision tasks such as image classification, it has triggered the interest to overcome the weakness of traditional methods and improve 3D reconstruction using deep models.

There are already several works that approach two-view stereo using deep models with successful results  (\eg, \cite{kendall2017end,mayer2016large}). However, rigid two-view stereo is a simpler problem than unstructured multi-view stereo, where camera motion can be arbitrary and varying. Yet, unstructured multi-view stereo is a highly relevant problem that appears in the context of depth estimation from moving monocular cameras. In fact, there are already robust and accurate approaches for real-time tracking of motion (\eg, commercially deployed solutions such as ARCore by Google and ARKit by Apple) but depth estimation remains a challenge. For example, some smartphones nowadays contain stereo cameras but the small baseline due to the size restriction of the device is a limitation for long-range depth estimation. Thus, multi-view depth estimation is helpful also in such cases since the additional baseline arising from motion can alleviate some of the problems.

Recently, various deep learning based multi-view depth estimation methods have been proposed, \eg \cite{huang2018deepmvs,zhou2018deeptam,wang2018mvdepthnet}. They typically  discretize the depth space and utilize a plane-sweep approach to compute a matching cost volume from which the disparity map is inferred via CNNs. The benefit is that the cost volume based approach force the network to learn disparity estimation via matching instead of just learning the single-view cues, which is beneficial for generalization. However, these methods have also problems: the depth range must be approximately known in advance and discretization poses an inherent trade-off between depth resolution and computational complexity. In addition, manually specified features and metrics are often used in the construction of the cost volume \cite{wang2018mvdepthnet,zhou2018deeptam} or the used networks are large and complex hampering computation speed \cite{huang2018deepmvs}.

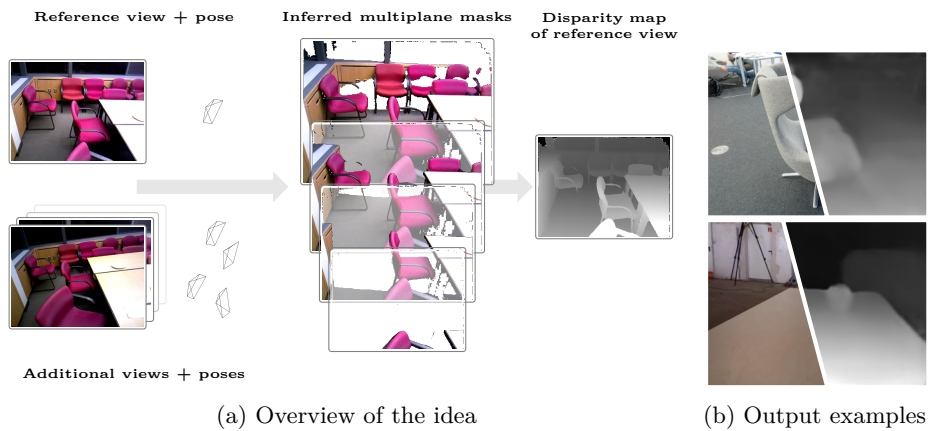
\begin{figure}[!t]
  \centering
  \begin{subfigure}[b]{.74\textwidth}

    \begin{tikzpicture}\tiny

      \tikzstyle{frame} = [rounded corners=1pt,minimum width=1.75cm,minimum height=1.3125cm,inner sep=1pt,draw=black!50,fill=white]

      \node [frame] (ref) at (0,0.5) {\includegraphics[width=1.75cm]{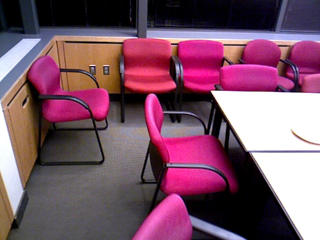}};

      \node [frame,draw=black!10] (n4) at (0.3,-1.4) {};            
      \node [frame,draw=black!25] (n3) at (0.2,-1.5) {};      
      \node [frame] (n2) at (0.1,-1.6) {\includegraphics[width=1.75cm]{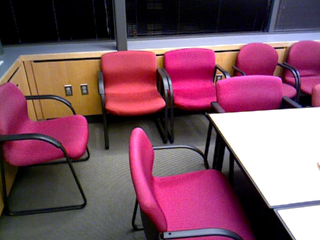}};
      \node [frame] (n1) at (0,-1.7) {\includegraphics[width=1.75cm]{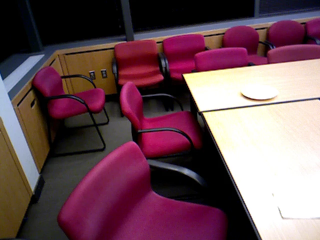}};

      \node [] (pose) at (1.8,0.5) {\includegraphics[scale=.08]{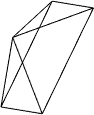}};
      \node [] (poses) at (1.8,-1.6) {\includegraphics[scale=.08]{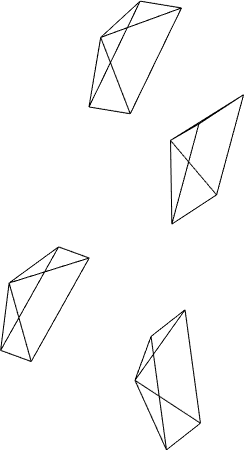}};

      \node at (0.75,1.75) {\bf Reference view + pose};
      \node at (0.75,-3) {\bf Additional views + poses};
      \node at (4.25,1.75) {\bf Inferred multiplane masks};      
      \node[text width=2.5cm,align=center,text centered,text depth = 0cm] at (7,1.75) {\bf Disparity map of reference view};

      \node[draw=black!10, single arrow, minimum height=20mm, minimum width=3mm,
            single arrow head extend=1mm, fill=black!10,
            anchor=center, rotate=0] at (1.75,-.5) {};
      \node[draw=black!10, single arrow, minimum height=20mm, minimum width=3mm,
            single arrow head extend=1mm, fill=black!10,
            anchor=center, rotate=0] at (5,-.5) {};

      \tikzstyle{mframe} = [rounded corners=1pt,minimum width=1.75cm,minimum height=1.3125cm,inner sep=1pt,draw=black!50]
      \tikzstyle{mframebg} = [rounded corners=1pt,minimum width=1.75cm,minimum height=1.3125cm,inner sep=1pt,fill=white,fill opacity=.5,draw=white,thick]
      \node [mframe] (ref) at (4.25,0.5) {\includegraphics[width=2.5cm]{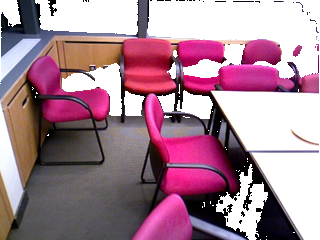}};      
      \node [mframebg,minimum width=2.25cm,minimum height=1.6875cm] (ref) at (4.25,-0.5) {};
      \node [mframe] (ref) at (4.25,-0.5) {\includegraphics[width=2.25cm]{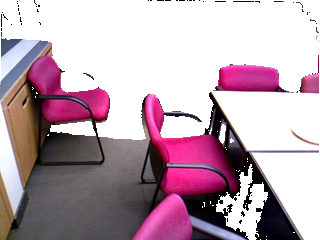}};
      \node [mframebg,minimum width=2cm,minimum height=1.5cm] (ref) at (4.25,-1.25) {};
      \node [mframe] (ref) at (4.25,-1.25) {\includegraphics[width=2.0cm]{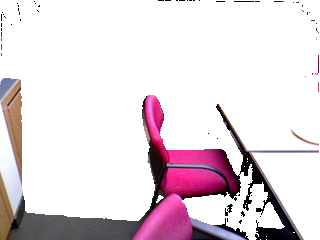}};      
      \node [mframebg] (ref) at (4.25,-2.0) {};
      \node [mframe] (ref) at (4.25,-2.0) {\includegraphics[width=1.75cm]{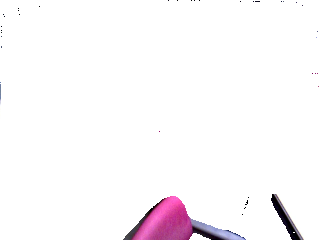}};

      \node [frame] (ref) at (7,-0.5) {\includegraphics[width=1.75cm]{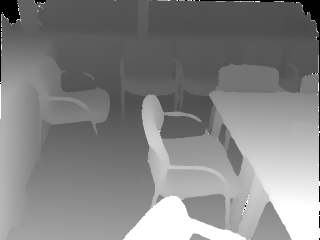}};
          
    \end{tikzpicture}
    \caption{Overview of the idea}
    \label{fig:sample-reconstructions}
  \end{subfigure}
  \hfill
  \begin{subfigure}[b]{.24\textwidth}
    \centering

    \setlength{\figurewidth}{\textwidth}
    \setlength{\figureheight}{0.75\figurewidth}

    \newcommand{\splitfig}[2]{\tikz{%
      \draw[draw=white,thick,inner sep=0,path picture={
        \node [anchor=west,text width=\figurewidth] at (path picture bounding box.west){
          \includegraphics[width=\figurewidth]{#1}
        };}] (0,0) -- (.54\figurewidth,0) -- 
         (.34\figurewidth,\figureheight) -- (0,\figureheight) -- (0,0);
      \draw[draw=white,thick,inner sep=0,path picture={
        \node [anchor=east,text width=\figurewidth] at (path picture bounding box.east){
          \includegraphics[width=\figurewidth]{#2}
        };}] (.55\figurewidth,0) -- (\figurewidth,0) -- (\figurewidth,\figureheight) --
         (.35\figurewidth,\figureheight) -- (.55\figurewidth,0);}}

    \splitfig{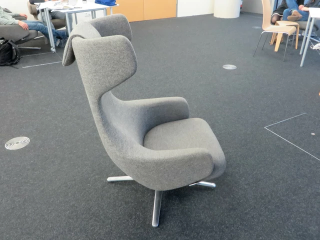}{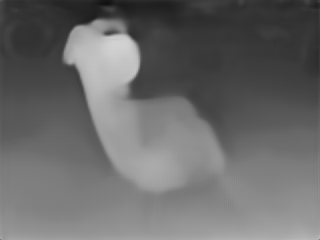}
    \splitfig{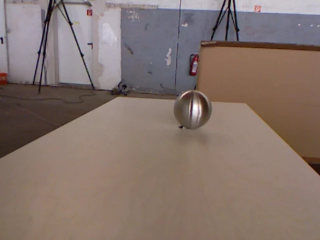}{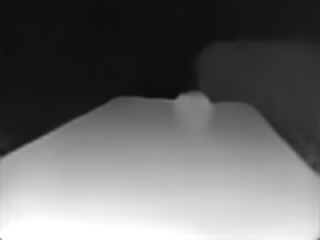}
    
    \caption{Output examples}
    \label{fig:sample-reconstructions}
  \end{subfigure}
  \caption{(a)~Overall idea of our method: To estimate a depth map with an arbitrary number of images and known camera poses, we back-project images onto a set of planes to generate the multiplane masks representation via a convolutional neural network.
The inferred masks are then passed through a second network to reconstruct the final disparity map. (b)~Our method performs well in textureless and varying depth cases.}
  \label{fig:teaser}
\end{figure}

In this paper, we propose our own plane-sweep based approach, which aims at avoiding some of the shortcomings of the previous methods. In particular, our method does not use manually specified features or cost metrics but instead infers a set of masking planes to regularize the learning of features. In addition, we propose selecting intermediate depth planes by depth histogram matching if the depth range of interest is approximately known {\it a~priori}. In comparison to recent approaches like \cite{huang2018deepmvs}, our architecture is relatively simple, lightweight, and more accurate.

In summary, the contributions of this paper are: {\em (i)}~We propose a CNN-based approach for multi-view stereo depth estimation that does not require constructing an explicit cost-volume metric; {\em (ii)}~We propose a way of selecting intermediate depth-planes by depth histogram matching; {\em (iii)}~We demonstrate that the current state-of-the-art in CNN-based MVS can be matched with a relatively simple and lightweight architecture.
This paper is structured as follows. Sec.~\ref{sec:related} goes through the background and covers related approaches. Sec.~\ref{sec:methods} presents our MaskMVS method in detail. Experiments and ablation studies are presented in Sec.~\ref{sec:experiments}, and this paper is concluded with a discussion in Sec.~\ref{sec:discussion}.

\section{Related Work}
\label{sec:related}
Multi-view stereo reconstruction has been under active research for long and only recently it has gotten a boost from CNN-based methods. Conventional MVS algorithms typically seek to design photometric error measures and solve an optimization task subject to penalizing visual inconsistency (see review in \cite{Furukawa+Hernandez:2015}). The most prominent traditional method is COLMAP~\cite{Schonberger+Zheng+Frahm+Pollefeys:2016} which jointly estimates depths and surface normals by leveraging photometric and geometric priors. While conventional MVS methods deliver impressive results in the best case, they fail in poorly textured regions where the photometric consistency is not reliable. They also cannot use visual cues for depth such as shadows or lighting, and typically require many frames as input.

The success of convolutional neural networks in computer vision has spawned a number of methods that leverage the capability of CNNs to learn visual cues. The extreme case is purely monocular depth estimation \cite{eigen2014depth,liu2016learning}, while left--right stereo reconstruction \cite{kendall2017end,mayer2016large} relies on the 1D correlation layer along the disparity line. Other approaches use nearby images as the supervision by warping and computing image reconstruction error \cite{godard2017unsupervised,zhou2017unsupervised,yin2018geonet}, but the CNN-based prediction still only utilize single view information.

Unlike left--right two-view stereo, images collected from monocular videos are more unstructured and they also suffer from dynamic moving objects, which makes the task more challenging. DeMoN~\cite{ummenhofer2017demon} can learn depth and motion for unconstrained image pairs, but it cannot handle multiple images as input. Current attempts on learned MVS mainly employ plane-sweeping approaches and regard the depth estimation problem as a multi-class classification problem \cite{huang2018deepmvs,zhou2018deeptam,yao2018mvsnet}.  In practice, these methods employ classical plane-sweep stereo approaches with a defined cost metric to build a cost volume, and the CNN is used to infer depth from the cost volume and refine the depth map. For example, DeepTAM~\cite{zhou2018deeptam} computes the sum of absolute difference (SAD) of $3{\times}3$ patches between warped image pairs. To increase the density of sampled planes, an adaptive narrow band strategy is used. DeepMVS~\cite{huang2018deepmvs} proposed a patch matching network to extract features that can aid in the comparison of patches. To do the feature aggregation, it considers both an intra-volume feature aggregation network and inter-volume aggregation network. Semantic features from pre-trained VGG-19 on ImageNet also aids in intra-volume feature aggreation. The Dense-CRF is used to refine the final depth map. MVDepthNet~\cite{wang2018mvdepthnet} computes the absolute difference directly without a supporting window to generate the cost volume, as the pixel-wise cost matching enable the volume to preserve detail information. MVSNet~\cite{yao2018mvsnet} proposes a variance-based cost metric and employ four-scale 3D CNN to obtain smooth cost volume automatically. DPSNet~\cite{im2018dpsnet} concatenate warped image features firstly and use a series of 3D convolutions to learn the cost volume generation.

Attempts on learned MVS without cost metrics have shown promise in reconstructing 3D objects. \cite{Hartmann+Galliani+Havlena+Gool+Schindler:2017} propose a CNN to learn multi-patch similarity directly, but it still matches patches explicitly. SurfaceNet~\cite{Ji+Gall+Zheng+Liu+Fang:2017}, and LSM~\cite{kar2017learning} use 3D grid to fuse information. However, due to the volumetric structure, SurfaceNet and LSM are limited to small-scale reconstructions (see discussion in \cite{yao2018mvsnet}).

\section{Methods}
\label{sec:methods}
\begin{figure}[!t]
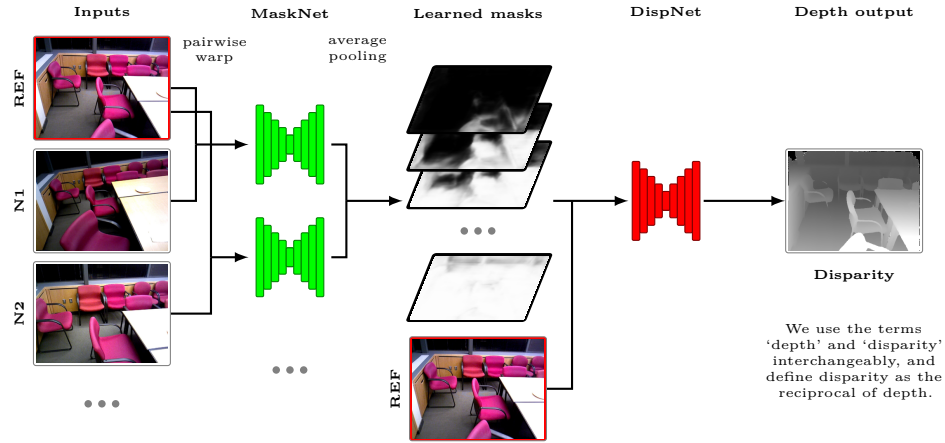

  \centering

\begin{tikzpicture}\tiny

  \newcommand{\drawnet}[3]{%
    \tikzstyle{block} = [rounded corners=0.5pt,minimum width=1mm,minimum height=1mm,inner sep=0,draw=#3!50!black,fill=#3]
    \foreach \j in {0,...,4} 
      \node[block,minimum height={0.25cm+0.2cm*\j}] at (#1-.1*\j,#2) {};  
    \foreach \j in {0,...,4} 
      \node[block,minimum height={0.25cm+0.2cm*\j}] at (#1+.1*\j,#2) {};}

  \newcommand{\drawmask}[3]{%
    \node at (#1,#2) {
      \pgfdeclareimage[width=1.5cm]{themask}{#3}
      \pgflowlevel{\pgftransformcm{1}{0}{0.3}{.75}{\pgfpoint{0}{0}}}
      \tikz\node[inner sep=1pt,fill=black,draw=black,thick,rounded corners=1pt]{\pgfuseimage{themask}};
    };}

  \newcommand{\drawdots}[2]{%
    \node at (#1,#2) {
      \tikz\node[circle,fill=black!50,inner sep=1pt]{}; 
      \tikz\node[circle,fill=black!50,inner sep=1pt]{};
      \tikz\node[circle,fill=black!50,inner sep=1pt]{};
    };}

  \tikzstyle{frame} = [rounded corners=1pt,minimum width=1.75cm,minimum height=1.3125cm,inner sep=1pt,draw=black!50]
  \node [frame,fill=red] (ref) at (0,0) {\includegraphics[width=1.75cm]{fig/ref}};
  \node [frame] (n1) at (0,-1.5) {\includegraphics[width=1.75cm]{fig/n1}};
  \node [frame] (n2) at (0,-3) {\includegraphics[width=1.75cm]{fig/n2}};

  \drawnet{2.5}{-0.75}{green}
  \drawnet{2.5}{-2.25}{green}

  \coordinate (d1) at (1.25,-0.75) {};
  \coordinate (d2) at (1.45,-2.25) {};
  \coordinate (p1) at (1.95,-0.75) {};
  \coordinate (p2) at (1.95,-2.25) {};
  \draw[black,thick] (ref) -| (d1);
  \draw[black,thick] (n1) -| (d1);
  \draw[black,thick] ([yshift=-.5cm]ref) -| (d2);
  \draw[black,thick] (n2) -| (d2);
  \draw[black,thick,-latex] (d1) -> (p1);
  \draw[black,thick,-latex] (d2) -> (p2);

  \node [frame,fill=red] (ref2) at (5,-4) {\includegraphics[width=1.75cm]{fig/ref}};

  \drawmask{4.75}{-2.5}{fig/mask4}

  \drawmask{4.75}{-1.0}{fig/mask3}
  \drawmask{4.75}{-0.5}{fig/mask2}
  \drawmask{4.75}{0}{fig/mask1}

  \drawnet{7.5}{-1.5}{red}

  \node [frame] (disp) at (10,-1.5) {\includegraphics[width=1.75cm]{fig/pred}};

  \coordinate (p3) at (3.05,-0.75) {};
  \coordinate (p4) at (3.05,-2.25) {};  
  \coordinate (m1) at (3.25,-1.5) {};
  \coordinate (m2) at (4,-1.5) {};
  \coordinate (m3) at (6.25,-1.5) {};
  \coordinate (m4) at (7,-1.5) {};
  \coordinate (m5) at (6,-1.5) {};
  \coordinate (m6) at (8,-1.5) {};
  \draw[black,thick] (p3) -| (m1);
  \draw[black,thick] (p4) -| (m1);
  \draw[black,thick,-latex] (m1) -> (m2);
  \draw[black,thick,-latex] (m5) -> (m4);
  \draw[black,thick] (ref2) -| (m3);
  \draw[black,thick,-latex] (m6) -> (disp);

  \drawdots{0}{-4.2}
  \drawdots{2.5}{-3.75}
  \drawdots{5}{-1.9}

  \node at (0,1) {\bf Inputs};
  \node at (2.5,1) {\bf MaskNet};
  \node at (5,1) {\bf Learned masks};  
  \node at (7.5,1) {\bf DispNet};    
  \node at (10,1) {\bf Depth output};
  \node [text width=1cm,align=center] at (1.5,.5) {pairwise warp};
  \node [text width=1cm,align=center] at (3.4,.5) {average pooling};
  \node [rotate=90] at (-1.1,0) {\bf REF};
  \node [rotate=90] at (-1.1,-1.5) {\bf N1};
  \node [rotate=90] at (-1.1,-3) {\bf N2};
  \node [rotate=90] at (3.9,-4) {\bf REF};
  \node at (10,-2.5) {\bf Disparity};
  \node[text width=2.5cm,align=justify,text centered,text depth = 0cm] at (10,-3.2) {We use the terms `depth' and `disparity' interchangeably, and define disparity as the reciprocal of depth.};  
    
\end{tikzpicture}

  \caption{Overview of our MaskMVS architecture. The MaskNet (left, green) will generate multi-plane mask maps to represent the probability of real surfaces being hit by a ray before each plane. Given the mask-based representation and the reference image, the DispNet (right, red) will learn to predict the disparity map for the reference image.}
  \label{fig:architecture}
\end{figure}

The overall architecture of MaskMVS is shown in Fig.~\ref{fig:architecture}. The estimation scheme consists of two parts, MaskNet and DispNet. Given (an arbitrary number of) image pairs and known camera poses, we back-project images onto virtual planes to construct the warped volume to feed the MaskNet. The MaskNet will generate multi-plane mask maps to represent the probability of a surface being hit by each ray before each plane. Given the mask-based representation and the reference image, the DispNet learns to predict the disparity map for the reference image.
Our methods will be introduced as follows. Sec.~\ref{sec:hist} presents our novel idea of depth plane sampling. Sec.~\ref{sec:masknet} and Sec.~\ref{sec:dispnet} explain the details of MaskNet and DispNet, respectively.

\subsection{Histogram-Based Depth Plane Sampling}
\label{sec:hist}
The selection of virtual planes is important for plane-sweeping methods. Current methods generally uniformly sample planes in the depth domain \cite{zhou2018deeptam,yao2018mvsnet} or inverse-depth domain \cite{wang2018mvdepthnet,im2018dpsnet} between predefined minimum and maximum values. One principle of plane selection is to achieve higher sampling density for close by depths and lower density for distant depths, so uniform sampling in the inverse-depth domain generally produces more accurate predictions (see \cite{im2018dpsnet}). However, both of these sampling methods rely on a fixed depth range, and the ideal depth ranges for indoor scenes and outdoor scenes are typically different. Some methods, such as \cite{huang2018deepmvs}, pre-run traditional methods like COLMAP~\cite{Schonberger+Zheng+Frahm+Pollefeys:2016} to obtain the depth range for each input firstly to deal with different scenes.

We propose the idea of selecting planes according to the cumulative histogram of depth. This allows us to sample reasonable numbers of depth planes in both nearby and far away depths when the training set is a mixed data set.
There will be more pixels covering areas closer to the camera in general, so the histogram of depth distribution is naturally in accordance with the selection principle mentioned above.  
We define the depth density and cumulative depth density functions as
\begin{equation}
  p(d_i) = \frac{n_i}{N} \qquad \text{and} \qquad P(d_i) = \sum_{j=1}^i p(d_i), 
\end{equation}
where $n_i$ is the frequency of the depth value $d_i$, and $N$ is the total number of pixels in the training data set. These discrete density functions, $p : [0,d_\mathrm{max}] \to [0,1]$ and $P : [0,d_\mathrm{max}] \to [0,1]$, can be seen as the normalized histogram and normalized cumulative histograms of the depths (we use a binning of 200 points in the experiments). Based on the cumulative density function, we can choose a set of depths covering the entire range by considering the inverse cumulative density function $P^{-1} : [0,1] \to[0,d_\mathrm{max}]$. By choosing to cover the quantiles $\theta_1, \theta_2, \ldots, \theta_D$ of $P$ uniformly, we find a set of depths $\{d_i\}_{i=1}^D$ such that $d_i = P^{-1}(\theta_i)$, where $D$ is the number of planes.

\subsection{MaskNet: Mask-Based Multiplane Representation}
\label{sec:masknet}
Similar to traditional plane-sweep stereo, we firstly construct a warped volume from each image pair by warping the neighbour image via the fronto-parallel planes at fixed depths to the reference frame using the planar homography:
\begin{equation}
  \mathbf{H} = \mathbf{K} \left(\mathbf{R} + \mathbf{t}\begin{pmatrix} 0 & 0 & \frac{1}{d_i} \end{pmatrix}\right)\mathbf{K}^{-1},
\end{equation}
where $\mathbf{K}$ is the known intrinsics matrix and the relative pose $(\mathbf{R}, \mathbf{t})$ is given in terms of a rotation matrix and translation vector with respect to the neighbour frame. $d_i$ denotes the depth value of the $i$\textsuperscript{th} virtual plane. The input of the MaskNet has size $3\,(1+D){\times}H {\times} W$ consisting of the reference image and concatenated warped neighbour images from $D$ planes ($D\!+\!1$ RGB images with height $H$ and width $W$). The output of the MaskNet is a set of  mask maps of size  $D {\times} H {\times} W$. 

Traditional plane-sweep based methods need to design a cost metric based on photo-consistency of warped images to select an optimal depth plane for each pixel. In that case, the predicted depth maps will be noisy and the accuracy will be limited by the density of the chosen planes.  Instead of using a distance metric, we propose a novel mask-based multiplane representation to roughly encode the near--far relationship. In our method, the intuition is that given a reference image and warped neighbour image on two successive planes, if the relative position of a warped pixel flips, it tells that the surface will be hit by the ray between the two planes. To enable the network to learn this, we assign a supervised binary classification task to the MaskNet, where the ground truth for masks can be obtained from ground truth depth maps directly. The MaskNet will predict whether the ray will hit a surface in front of the plane (including on the plane) or behind the plane for each pixel.

For this purpose, the MaskNet consists of an encoder--decoder architecture that shares a similar architecture with \cite{wang2018mvdepthnet}. The encoder includes five convolutional layers, and convolutional filter sizes decrease towards deeper layers: $7{\times}7$ for the first layer, $5{\times}5$ for the second layer, and $3{\times}3$ for the following three layers. There are four skip connections between the encoder and the decoder, and mask maps are predicted in four scales. All layers are followed by batch normalization and ReLU except for the mask prediction layers that are followed by a sigmoid function. Each pixel on each plane has a value in range $[0,1]$, representing the predicted probability that the true surface is located in front of the sampled plane. To support arbitrary length of sequence, we deal with neighbour images separately and then use average pooling for predicted masks with the finest resolution to fuse information from different neighbours.  We compute the pixel-wise cross-entropy between the averaged mask map and the ground truth mask map as loss function to train the MaskNet.

\subsection{DispNet: Continuous-Disparity Prediction}
\label{sec:dispnet}
To regress continuous depth values, given prediction results from MaskNet and the reference image, we concatenate them into an input of size $(3+D) {\times} H {\times} W$ to feed the DispNet. The encoder--decoder architecture of our DispNet is similar to the DispNet in \cite{mayer2016large}. There are six convolutional layers for the encoder ($3 {\times} 3$ filters except for the first two layers, which have a $7 {\times} 7$ filter and a $5 {\times} 5$ filter, respectively) to extract features, and the decoder will gradually upsample the feature maps, considering also features from the encoder with six skip connections. All convolutional layers are followed by batch normalization and LeakyReLU, and all deconvolution layers are followed by LeakyReLU. Unlike \cite{wang2018mvdepthnet,zhou2017unsupervised} that use scale and sigmoid functions to constrain the range of the output, our inverse depth prediction layer consists of a convolutional layer and a ReLU layer to only ensure that predicted inverse depth maps have values larger than zero for all pixels. The DispNet will generate depth maps in six resolutions and the finest resolution is the same as in the reference image. During training, the loss function is defined as the sum of the average L1 loss at different resolutions:
\begin{equation}
L = \sum_{s=0}^{6} w_s\frac{1}{n_s}\sum_i|\hat{d}_{s,i} - d_{s,i}|,
\end{equation}
where $\hat{d}_{s,i}$ is the estimated inverse depth at scale $s$ and $d_{s,i}$ is the corresponding ground truth inverse depth. $n_s$ is the number of valid pixels and $w_s$ is the loss weight for scale $s$. We assign the highest weight for the loss with the finest resolution as $0.5$ and others $0.1$.

\begin{figure}[!t]
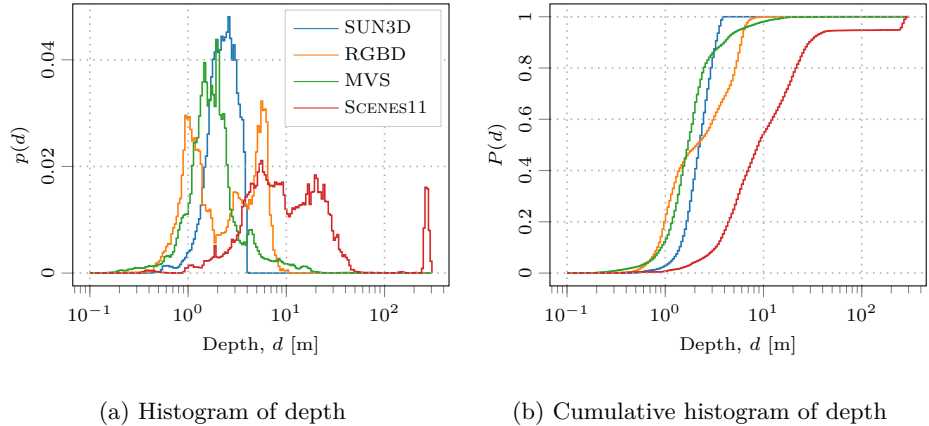

  \centering
  \setlength{\figurewidth}{.41\textwidth}
  \setlength{\figureheight}{.75\figurewidth}
  \pgfplotsset{yticklabel style={rotate=90}, ylabel style={yshift=-15pt},clip=true,scale only axis,axis on top,clip marker paths,legend style={row sep=0pt},xlabel near ticks,legend style={fill=white},every axis plot/.append style={const plot}}
  \begin{subfigure}{.48\textwidth}
    \centering\scriptsize
 	\input{./fig/pdf.tex}
 	\caption{Histogram of depth}
 	\label{fig:pdf}
  \end{subfigure}  
  \hfill
  \begin{subfigure}{.48\textwidth}
    \centering\scriptsize
  	\input{./fig/cdf.tex}
 	\caption{Cumulative histogram of depth}
 	\label{fig:cdf}  	
  \end{subfigure}  	 
  \caption{Depth distributions (log-scale) for the four data sets used in training and evaluation. The typical depths in the data are highly depenedent on the type of environments covered (\eg, \textsc{sun3d} covering mostly indoor scenes and \textsc{scenes11} having high variability in the type of scenes covered.)}
  \label{fig:distribution}
\end{figure}

\section{Experiments}
\label{sec:experiments}
Similar to DeepMVS~\cite{huang2018deepmvs}, we train our networks with the same data sets as used in DeMoN~\cite{ummenhofer2017demon}. The training data set includes short sequences from real-world data sets \textsc{sun3d}~\cite{xiao2013sun3d},  RGBD~\cite{sturm2012benchmark}, MVS (includes \textsc{Citywall} and \textsc{Achteck-Turm}~\cite{fuhrmann2014mve}), and a synthesized data set \textsc{scenes11}~\cite{ummenhofer2017demon}. \textsc{sun3d} consists of a variety of indoors scenes, RGBD provides scenes of an office and an industrial hall. MVS data sets include both indoor and outdoor scenes, and the ground truth depth maps of outdoor scenes are often sparse. \textsc{scenes11} provides perfect ground truth but lack realism. Combined, there are 92,558 training samples and each training sample consists of a sequence of three frames, the ground truth depth map for the reference frame, and provided ground truth camera poses.  The resolution of the input images is $320{\times}240$.

Our training procedure consists of two stages as we pre-train the MaskNet first and then employed the pre-trained parameters of MaskNet to predict masks to train the DispNet. For both networks we use the Adam solver \cite{kingma2014adam} with $\beta_1 = 0.9$ and $\beta_2 = 0.999$. The learning rate for MaskNet and DispNet is $2\cdot10^{-4}$ and $10^{-4}$, respectively. Our framework is trained using only $D = 16$ sampled depth planes as sparse sampling can present the effect of plane selection more clearly and provide speed-up. The whole framework is implemented on PyTorch for 500k iterations with a mini-batch size of four. 

{~\\[-6pt]\noindent\bf Error Metrics~~}
In our evaluations, we use three common measures: {\em (i)}~L1-rel, {\em (ii)}~L1-inv, and {\em (iii)}~sc-inv (see, \eg, \cite{eigen2014depth}). The two L1 metrics are the mean absolute relative difference and mean absolute difference in inverse depth, respectively. They are given in terms of
\begin{equation}
  \text{L1-rel} = \frac{1}{n}\sum_i \frac{|d_i - \hat{d}_i|}{\hat{d}_i} 
  \qquad \text{and} \qquad
  \text{L1-inv} = \frac{1}{n}\sum_i \left|\frac{1}{d_i} - \frac{1}{\hat{d}_i}\right|,
\end{equation}
where $d_i$ [meters] is the predicted depth value, $\hat{d}_i$ [meters] is the ground truth value, $n$ is the number of pixels for which the depth is available. The third, scale-invariant metric, is given as:
\begin{equation}
  \text{sc-inv} = \sqrt{\frac{1}{n} \sum_i z_i^2 - \frac{1}{n^2}\bigg(\sum_i z_i\bigg)^2}
\end{equation}
where $z_i = \log d_i - \log \hat{d}_i$. The L1-rel metric normalizes the error, L1-inv metric gives more importance to close depth values and sc-inv is a scale-invariant metric.

\begin{figure}[!t]
  \centering
  \setlength{\figurewidth}{.45\textwidth}
  \setlength{\figureheight}{.30\figurewidth}
  \pgfplotsset{yticklabel style={rotate=90}, ylabel style={yshift=-15pt},clip=true,scale only axis,axis on top,clip marker paths,legend style={row sep=0pt},xlabel near ticks,legend style={fill=white},every axis plot/.append style={const plot}}
  \begin{subfigure}[b]{.48\textwidth}
    \centering\scriptsize
\begin{tikzpicture}

\definecolor{color0}{rgb}{0.12156862745098,0.466666666666667,0.705882352941177}

\begin{axis}[
height=\figureheight,
tick align=outside,
tick pos=left,
width=\figurewidth,
xmajorgrids,
xmin=0.0669698609155416, xmax=453.453400756686,
xmode=log,
y grid style={white!69.01960784313725!black},
ymajorgrids,
ymin=-0.05, ymax=1.05,
xticklabels={,,}
]
\addplot [semithick, color0, forget plot]
table [row sep=\\]{%
0.1	0 \\
0.104132894355703	1.01255060525415e-06 \\
0.108436596868961	3.11865586418278e-06 \\
0.112918166860475	6.48032387362656e-06 \\
0.117584955405216	1.15835789241075e-05 \\
0.122444617390314	1.78613926766832e-05 \\
0.127505124071301	2.27216355819031e-05 \\
0.132774776147277	3.21586072228718e-05 \\
0.138262217376466	4.21626072027828e-05 \\
0.143976448754488	5.92139593952627e-05 \\
0.149926843278605	8.23406152192674e-05 \\
0.15612316132215	0.000105507773067482 \\
0.162575566644379	0.000135357764910375 \\
0.169294643061978	0.00017071603204585 \\
0.176291411809595	0.000207896890270782 \\
0.183577349617863	0.000262574622954506 \\
0.19116440753857	0.000347507367723224 \\
0.199065030547846	0.000449572468732842 \\
0.207292177959537	0.000559656970536073 \\
0.215859344682242	0.000659858978432024 \\
0.224780583354873	0.000765326249475296 \\
0.234070527397063	0.000878691415239551 \\
0.243744415012222	0.00102093452426565 \\
0.253818114182605	0.00116969845918959 \\
0.264308148697411	0.0013235251471398 \\
0.27523172525659	0.0015063917864487 \\
0.286606761694825	0.00171149403704898 \\
0.298451916371975	0.00190983244960617 \\
0.310786618778201	0.00211347662733488 \\
0.323631101403967	0.00231805235162043 \\
0.337006432927193	0.00261898239150196 \\
0.350934552771998	0.00295324559730846 \\
0.365438307095726	0.00339856535349924 \\
0.380541486263263	0.0038701304213782 \\
0.396268863870148	0.00436482214508117 \\
0.412646237378447	0.00483638721296013 \\
0.429700470432084	0.00536793577869435 \\
0.447459536921003	0.00608441658697218 \\
0.465952566866468	0.00686675368661575 \\
0.485209894202748	0.00770822374160616 \\
0.505263106533568	0.00863879824985893 \\
0.526145096944946	0.00979881672526229 \\
0.547890117959394	0.0113304817748181 \\
0.570533837719995	0.0130808577551089 \\
0.594113398496503	0.0150474335406335 \\
0.618667477609444	0.0171334092935297 \\
0.644236350872137	0.0191073159454363 \\
0.670861958654722	0.0216105840537698 \\
0.698587974678525	0.0243985408902766 \\
0.727459877653637	0.0275310889487635 \\
0.757525025877191	0.0312152340749686 \\
0.788832734914711	0.0357479371104009 \\
0.821434358491943	0.0407912086630266 \\
0.855383372729865	0.0471193664316955 \\
0.890735463861044	0.0547245935257352 \\
0.927548619571206	0.0641717716768049 \\
0.96588322411587	0.0736636640626025 \\
1.00580215736804	0.0838466829895225 \\
1.04737089795945	0.0938210759897358 \\
1.0906576306845	0.104513650883244 \\
1.13573335834311	0.116070579475421 \\
1.18267201820591	0.127432652829123 \\
1.23155060329283	0.139270624967295 \\
1.28244928866395	0.151878095063435 \\
1.3354515629299	0.165138255279722 \\
1.39064436519738	0.178002184193162 \\
1.44811822767453	0.192631636843946 \\
1.50796742417001	0.206828406370094 \\
1.57029012472938	0.221093988835374 \\
1.63518855666249	0.237025824576661 \\
1.7027691722259	0.254094471133478 \\
1.77314282323548	0.271161497609328 \\
1.84642494289554	0.288473318319503 \\
1.92273573514278	0.307866983576105 \\
2.00220037181558	0.326395039571288 \\
2.08494919797222	0.343001153011625 \\
2.1711179456945	0.360145538353716 \\
2.26084795672778	0.377433543873656 \\
2.35428641432242	0.394596843661052 \\
2.45158658465704	0.410915230721401 \\
2.55290806823952	0.427690075594598 \\
2.65841706169809	0.443306319563239 \\
2.76828663039207	0.458620742447466 \\
2.88269699228923	0.474913856244707 \\
3.00183581357559	0.49063180079614 \\
3.12589851648234	0.505193129042203 \\
3.25508859983506	0.518995530358616 \\
3.38961797285079	0.532688009181323 \\
3.52970730273065	0.54751065648759 \\
3.67558637661806	0.561582274758928 \\
3.82749447851631	0.57366354255653 \\
3.98568078178378	0.583455716949822 \\
4.15040475785047	0.593929540410571 \\
4.32193660182653	0.604887282056583 \\
4.5005576757005	0.615998931394593 \\
4.68656096985471	0.627354969946688 \\
4.88025158365443	0.639767422796257 \\
5.08194722589941	0.654182781747066 \\
5.29197873595844	0.66881280142619 \\
5.5106906264419	0.684909682916133 \\
5.7384416483024	0.699724310821606 \\
5.97560537929042	0.714351981383326 \\
6.22257083673023	0.728045796772831 \\
6.4797431156211	0.738879318710591 \\
6.74754405311069	0.748222892683731 \\
7.02641292043031	0.757122280957358 \\
7.3168071434272	0.764908714107714 \\
7.61920305287561	0.772829654480472 \\
7.93409666579749	0.780328766271081 \\
8.26200449907429	0.787768299584077 \\
8.6034644166845	0.794872152120419 \\
8.95903651195662	0.802177218713037 \\
9.32930402628469	0.808416150522371 \\
9.71487430581342	0.813566023402718 \\
10.1163797976621	0.818395403765489 \\
10.5344790873212	0.822879747384015 \\
10.9698579789238	0.827566236605373 \\
11.4232306201635	0.832703067833924 \\
11.8953406737032	0.837695185329972 \\
12.3869625369984	0.842811441530225 \\
12.8989026125331	0.847832801487753 \\
13.4320006305542	0.853037554610904 \\
13.9871310264724	0.858324931863445 \\
14.5652043751903	0.863782782135886 \\
15.1671688847092	0.869104748117101 \\
15.7940119514654	0.87519779163523 \\
16.4467617799466	0.882002941743022 \\
17.126489069246	0.887267759368078 \\
17.8343087693191	0.893371252408453 \\
18.571381909825	0.899496373529756 \\
19.3389175045523	0.906853647231581 \\
20.1381745345521	0.913953530569551 \\
20.9704640132323	0.919998984209233 \\
21.8371511368	0.926078581055324 \\
22.7396575235793	0.931623389173744 \\
23.6794635458776	0.938250330375011 \\
24.658110758226	0.943613932437115 \\
25.6772044259759	0.948962143730019 \\
26.7384161583995	0.953397884919493 \\
27.8434866506145	0.957727551307559 \\
28.9942285388288	0.960898333276901 \\
30.1925293735898	0.963360329822564 \\
31.440354715915	0.966043264910294 \\
32.7397513613822	0.968927616564421 \\
34.0928506974681	0.970704399864496 \\
35.5018721996422	0.972303176768168 \\
36.9691270719503	0.973558172490345 \\
38.4970220380597	0.974937104406604 \\
40.0880632889846	0.976118426946742 \\
41.7448605939659	0.977117085357692 \\
43.4701315812502	0.977697317356527 \\
45.2667061957886	0.977951265048325 \\
47.1375313411672	0.978207359347405 \\
49.0856757133843	0.978339152934185 \\
51.1143348344017	0.978432712610111 \\
53.226836293728	0.97851379766258 \\
55.4266452066311	0.978579248933703 \\
57.7173698979317	0.978623112625923 \\
60.1027678207038	0.978661994569165 \\
62.5867517195873	0.978699863961801 \\
65.1733960488242	0.978731091022467 \\
67.8669436555464	0.978762925613496 \\
70.6718127392749	0.978809948463604 \\
73.5926040990498	0.978845549742885 \\
76.6341086800745	0.978869607945266 \\
79.801315432257	0.978893423135501 \\
83.099419493534	0.978928376382395 \\
86.5338307114045	0.978956768301366 \\
90.1101825166502	0.978979449434924 \\
93.834341163795	0.97899605526485 \\
97.712415353465	0.979013349629188 \\
101.75076625243	0.979024649693942 \\
105.956017927762	0.97902991495709 \\
110.335068212226	0.979032588090688 \\
114.895100018731	0.979035220722261 \\
119.643593122385	0.979036111766794 \\
124.588336429501	0.979041134017796 \\
129.73744075366	0.979056403280923 \\
135.099352119803	0.979102697094595 \\
140.682865618154	0.979179772446667 \\
146.497139830729	0.979273615636762 \\
152.55171185406	0.979336758292506 \\
158.856512942805	0.979367701839002 \\
165.421884799886	0.979379690438169 \\
172.258596539879	0.979386413774188 \\
179.377862353489	0.979396741790361 \\
186.791359902078	0.97940836587131 \\
194.511249472413	0.979432221563569 \\
202.550193923067	0.979470779490617 \\
210.92137945518	0.979498523377201 \\
219.638537241655	0.979510876494585 \\
228.715965950265	0.979513306616038 \\
238.168555197616	0.97985704729551 \\
248.011809972439	0.983221347936527 \\
258.261876068267	0.98962698607751 \\
268.935566567227	0.995933434761202 \\
280.050389418363	0.999120134026058 \\
291.62457615576	1 \\
303.677111803546	1 \\
};
\addplot [semithick, red, forget plot]
table [row sep=\\]{%
1.1162109375	0 \\
1.1162109375	1 \\
};
\addplot [semithick, red, forget plot]
table [row sep=\\]{%
1.353515625	0 \\
1.353515625	1 \\
};
\addplot [semithick, red, forget plot]
table [row sep=\\]{%
1.595703125	0 \\
1.595703125	1 \\
};
\addplot [semithick, red, forget plot]
table [row sep=\\]{%
1.8359375	0 \\
1.8359375	1 \\
};
\addplot [semithick, red, forget plot]
table [row sep=\\]{%
2.078125	0 \\
2.078125	1 \\
};
\addplot [semithick, red, forget plot]
table [row sep=\\]{%
2.375	0 \\
2.375	1 \\
};
\addplot [semithick, red, forget plot]
table [row sep=\\]{%
2.7265625	0 \\
2.7265625	1 \\
};
\addplot [semithick, red, forget plot]
table [row sep=\\]{%
3.15234375	0 \\
3.15234375	1 \\
};
\addplot [semithick, red, forget plot]
table [row sep=\\]{%
3.701171875	0 \\
3.701171875	1 \\
};
\addplot [semithick, red, forget plot]
table [row sep=\\]{%
4.5234375	0 \\
4.5234375	1 \\
};
\addplot [semithick, red, forget plot]
table [row sep=\\]{%
5.42578125	0 \\
5.42578125	1 \\
};
\addplot [semithick, red, forget plot]
table [row sep=\\]{%
6.2890625	0 \\
6.2890625	1 \\
};
\addplot [semithick, red, forget plot]
table [row sep=\\]{%
8.0234375	0 \\
8.0234375	1 \\
};
\addplot [semithick, red, forget plot]
table [row sep=\\]{%
11.765625	0 \\
11.765625	1 \\
};
\addplot [semithick, red, forget plot]
table [row sep=\\]{%
17.859375	0 \\
17.859375	1 \\
};
\addplot [semithick, red, forget plot]
table [row sep=\\]{%
25.7031269073486	0 \\
25.7031269073486	1 \\
};
\end{axis}

\end{tikzpicture}
 	\input{./fig/inverse_sampling.tex}
 	\caption{Two ways of depth plane sampling}
 	\label{fig:sampling}  	 	
  \end{subfigure} 
  \hfill
  \begin{subfigure}[b]{.48\textwidth}
    \centering\scriptsize
    \setlength{\figurewidth}{.25\textwidth}
    \setlength{\figureheight}{0.75\figurewidth}
  	\newcommand{\figg}[1]{\includegraphics[width=.95\figurewidth]{./fig/ablation/#1}}

   \newcommand{\figrow}[2]{%
     \node [draw=white,thick,minimum width=\figurewidth,inner sep=0] at
       ({0*\figurewidth},#2) {\figg{#1-ref}};%
     \node [draw=white,thick,minimum width=\figurewidth,inner sep=0] at
       ({1*\figurewidth},#2) {\figg{#1-gt}};%
     \node [draw=white,thick,minimum width=\figurewidth,inner sep=0] at
       ({2*\figurewidth},#2) {\figg{#1-inverse}};%
     \node [draw=white,thick,minimum width=\figurewidth,inner sep=0] at
       ({3*\figurewidth},#2) {\figg{#1-hist}};%
  }
  
  \begin{tikzpicture}

  \def\myarray{{"Ref","GT","Inverse","Histogram"}}
  \foreach \i in {0,...,3}
     \node[text width=.9\figurewidth,align=center,text centered,text depth = 0.3cm] at ({\figurewidth*\i},{-.8*\figureheight}) {\scriptsize \sc \vphantom{$^\dagger$}\pgfmathparse{\myarray[\i]}\pgfmathresult};

  \figrow{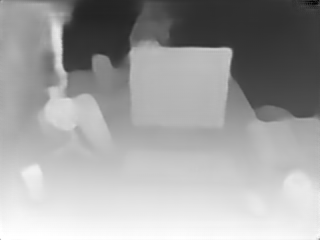}{0}  
  \figrow{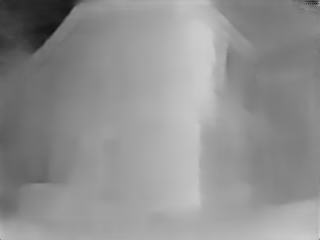}{1.03\figureheight}
  \figrow{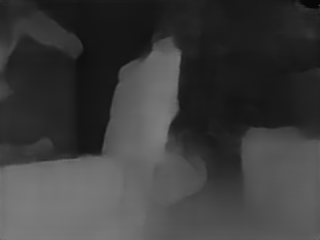}{2.06\figureheight}
  \figrow{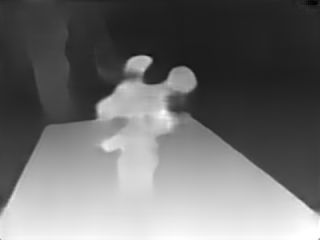}{3.09\figureheight}

  \end{tikzpicture}   
 	\caption{Comparison of disparity maps}
 	\label{fig:qualitative-comparison}  	
  \end{subfigure}   
  \caption{Setup and results for the ablation study. In (a), the upper figure shows the chosen depth plane depths for the histogram-based sampling and the bottom shows the depths for inverse depth based sampling. In (b), we show qualitative comparison between the two depth plane selection methods used in combination with our MaskMVS.}
  \label{fig:ablation}
\end{figure}

\subsection{Ablation Study: The Effect of Plane Selection}
\label{sec:ablation}
As the training set is a mixed data set that consists of both indoor scenes, outdoor scenes, and synthesized scenes, the depth ranges of image samples vary with type of data set. Fig.~\ref{fig:distribution} shows the distribution of depth for different sets separately. To examine the effects of plane selections, we conducted an ablation study for our method. We compare two options for plane selection: uniform sampling in the inverse-depth space and uniform sampling in the distribution space. Namely, to uniformly sample $D=16$ planes in the inverse-depth space from $d_\mathrm{min}=0.5$ meter to $d_\mathrm{max}=50$ meter, the $i$\textsuperscript{th} depth plane is given by: 
\begin{equation}
  \frac{1}{d_i} = \left(\frac{1}{d_\mathrm{min}} - \frac{1}{d_\mathrm{max}} \right)\frac{i}{D - 1} + \frac{1}{d_\mathrm{max}}.
\end{equation}
To uniformly sample planes in the distribution space of the whole data set, we set $\theta_\mathrm{min}=0.1$ and $\theta_\mathrm{max} = 1$, then the $i$\textsuperscript{th} depth plane is given by: 
\begin{equation}
  \theta_i = \theta_\mathrm{min} + (\theta_\mathrm{max}-\theta_\mathrm{min})\frac{i}{D}, \quad \text{such that} \quad d_i = P^{-1}(\theta_i), 
\end{equation}
where $i\in \{0,1,...,D-1\}$. Fig.~\ref{fig:sampling} shows the two sampling schemes. The curve is the cumulative histogram of depth of the whole mixed data set and the vertical lines present sampled depths. The higher slope of the curve corresponds to denser distribution of objects. It shows that using histogram-based sampling successfully gives planes within the range with higher slope, and the density is also higher in the closer areas than distant areas.

The evaluation results in Table~\ref{tbl:results} shows that selecting histogram-based planes perform much better in outdoor and synthesized scenes as it has more planes put in for distant depths. The performance of indoor scenes remains comparable with using inverse-depth sampled planes as it still samples densely-enough in close by depths. Fig.~\ref{fig:qualitative-comparison} shows comparison of the disparity maps from the two sampling approaches. Generally, using histogram-based sampling can provide good quality of prediction in both small-scale and large-scale depth scenes even with sparse depth planes. The last row in the figure shows that our methods with both sampling strategies give good predictions, but using histogram-based sampling failed to capture small objects like cans on the table in the close areas. It is mainly because our sampled planes started from the value farther than $1$~meter, while the office scenes in RGBD contain many objects within $1$ meter (see the first bump of RGBD in Fig.~\ref{fig:pdf}); conversely, inverse-space sampled planes are very dense within the range, so its prediction captures these details well.

\begin{table}[!tb]
  \caption{Comparison results between MVDepthNet, DeepMVS, COLMAP, and our method. We outperform other methods in most of the data sets and error metrics (smaller better).}
  \label{tbl:results}
  \centering
  {\noindent\scriptsize%
  \begin{tabular*}{\textwidth}{%
    >{\raggedright}p{0.13\textwidth} 
    P{0.166\textwidth} P{0.166\textwidth} P{0.166\textwidth} P{0.166\textwidth} P{0.166\textwidth}}
  \toprule
    & MVDepth-16 & DeepMVS &  COLMAP & Ours (hist)  & Ours (inv.) \\
\multicolumn{6}{l}{\bf scenes11 \hfill \rule[3pt]{.87\textwidth}{.1pt}} \\
~~L1-rel	 & 0.2352 & 0.3755 & 0.7205 & $\mathbf{0.1475}$ & 0.2144\\
~~L1-inv	 & 0.0292 & 0.0495 & 0.0936 & $\mathbf{0.0231}$ & 0.0308\\
~~sc-inv	& 0.3207 & 0.5810 & 0.7814 & $\mathbf{0.2483}$ & 0.2985\\
\multicolumn{6}{l}{\bf MVS \hfill \rule[3pt]{.87\textwidth}{.1pt}} \\
~~L1-rel	 & 0.3835 & 0.8217 & 0.9921 & $\mathbf{0.2669}$ & 0.4030\\
~~L1-inv	 & 0.1384 & $\mathbf{0.1065}$ & 0.1812 & 0.1377 & 0.1600\\
~~sc-inv	 & 0.3427 & 0.5325 & 0.6892 & $\mathbf{0.3001}$ & 0.3100\\
\multicolumn{6}{l}{\bf sun3d \hfill \rule[3pt]{.87\textwidth}{.1pt}} \\
~~L1-rel	& 0.1840 & 0.8604 & 1.8499 & 0.1797 & $\mathbf{0.1611}$\\
~~L1-inv	& 0.0865 & 0.1317 & 0.4511& $\mathbf{0.0818}$ & 0.0808\\
~~sc-inv	 & 0.2013 & 0.4992 &1.1219 & 0.1916 & $\mathbf{0.1769}$\\
\multicolumn{6}{l}{\bf RGBD \hfill \rule[3pt]{.87\textwidth}{.1pt}} \\
~~L1-rel	 & 0.1628 & 0.5066 & 2.2992 & 0.1748 & $\mathbf{0.1572}$\\
~~L1-inv	 & $\mathbf{0.0789}$ & 0.1717 & 0.5593 &0.0846 & 0.0802\\
~~sc-inv	 & 0.2360 & 0.5238 & 1.2970 & 0.2304 & $\mathbf{0.2062}$\\
  \bottomrule
  \end{tabular*}}
\end{table}

\subsection{Comparisons}
\label{sec.comparisons}
We provide both qualitative and quantitative comparisons to the state-of-the-art by evaluating using unstructured view pairs from the test sets in MVS, \textsc{sun3d}, RGBD, and \textsc{scenes11}. We compare our methods with two CNN-based multiview stereo methods (DeepMVS \cite{huang2018deepmvs} and MVDepthNet \cite{wang2018mvdepthnet}) and one traditional multiview stereo method (COLMAP \cite{Schonberger+Zheng+Frahm+Pollefeys:2016}).  The original MVDepthNet is trained with 64 planes and a larger data set (covering also the standard test samples in the sets). In order to make a fair comparison, we retrained the MVDepthNet with our training data set and 16 planes. The results are reported in Table~\ref{tbl:results}. Our predictions have significantly lower errors in \textsc{scenes11} and MVS, and comparable performance in \textsc{sun3d} and RGBD. The improvement of the outdoor scenes and synthesized scenes can be explained by the consideration of far depth planes. As mentioned in Sec.~\ref{sec:ablation}, almost half of the scenes of RGBD include objects closer than $1$~meter that is below our histogram-based sampling range, so the performance is slightly worse. We also evaluated the inference time for CNN-based models on a desktop workstation (NVIDIA GTX~1080~Ti, i7-7820X CPU and 63~GB memory; average over 100 predictions): DeepMVS 5.81~s, MVDepth-16 0.063~s, ours 0.089~s. The running time of our model is comparable to MVdepth-16 but our accuracy is better. Both models are significantly faster than DeepMVS.

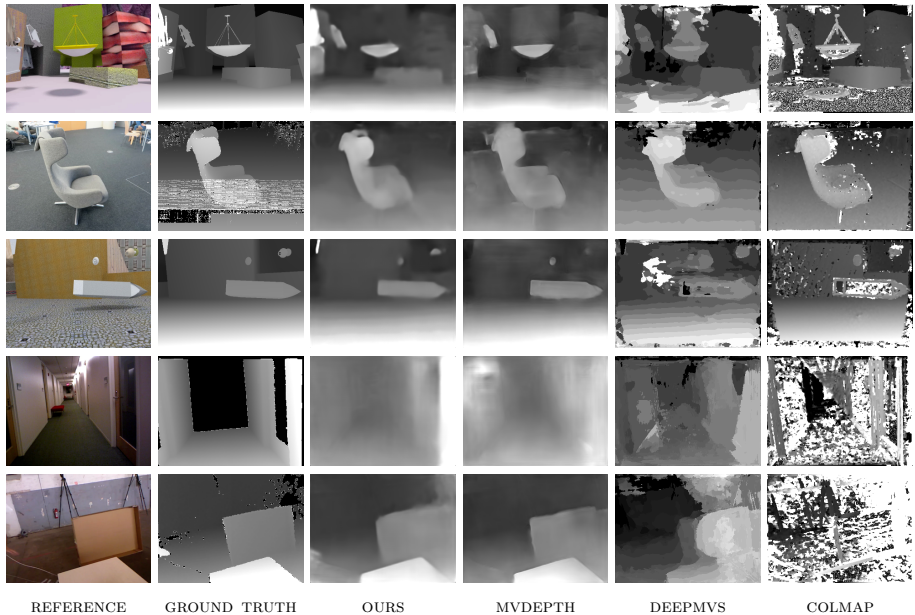
\begin{figure*}[!t]
  \centering
  \setlength{\figurewidth}{.166\textwidth}
  \setlength{\figureheight}{0.75\figurewidth}

  \newcommand{\figg}[1]{\includegraphics[width=.95\figurewidth]{./fig/comparison/#1}}

  \newcommand{\figrow}[2]{%
     \node [draw=white,thick,minimum width=\figurewidth,inner sep=0] at
       ({0*\figurewidth},#2) {\figg{#1-ref}};%
     \node [draw=white,thick,minimum width=\figurewidth,inner sep=0] at
       ({1*\figurewidth},#2) {\figg{#1-gt}};%
     \node [draw=white,thick,minimum width=\figurewidth,inner sep=0] at
       ({2*\figurewidth},#2) {\figg{#1-masknet}};%
     \node [draw=white,thick,minimum width=\figurewidth,inner sep=0] at
       ({3*\figurewidth},#2) {\figg{#1-mvdepth}};%
     \node [draw=white,thick,minimum width=\figurewidth,inner sep=0] at
       ({4*\figurewidth},#2) {\figg{#1-deepmvs}};%
     \node [draw=white,thick,minimum width=\figurewidth,inner sep=0] at
       ({5*\figurewidth},#2) {\figg{#1-colmap}};%
  }

\begin{tikzpicture}

  \def\myarray{{"reference","ground truth","ours","mvdepth","deepmvs","colmap"}}
  \foreach \i in {0,...,5}
     \node[text width=.9\figurewidth,align=center,text centered,text depth = 0cm] at ({\figurewidth*\i},{-.65*\figureheight}) {\scriptsize \sc \vphantom{$^\dagger$}\pgfmathparse{\myarray[\i]}\pgfmathresult};

  \figrow{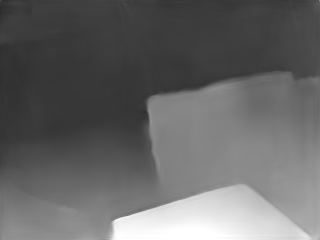}{0}  
  \figrow{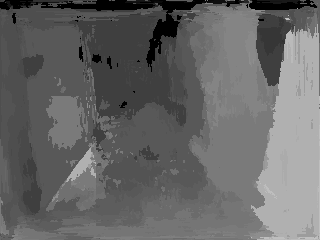}{1.03\figureheight}
  \figrow{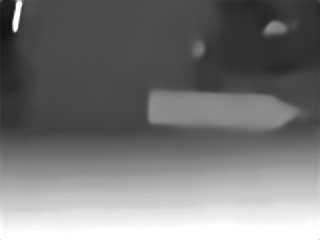}{2.06\figureheight}
  \figrow{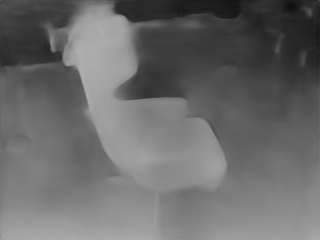}{3.09\figureheight}
  \figrow{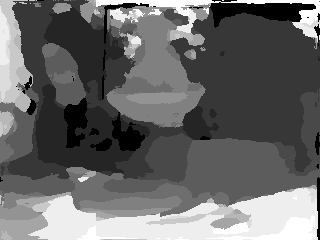}{4.12\figureheight}
        
\end{tikzpicture}   
  
  \caption{Qualitative comparisons between different algorithms on the MVS, \textsc{Scenes11}, \textsc{Sun3d}, and RGBD test sets. The traditional COLMAP method fails in low-texture environments. Our methods successfully captures small objects in close areas and provides better shape estimates for objects in far areas at the same time. Missing values in ground truth are shown in black.}
  \label{fig:frames}
\end{figure*}

Fig.~\ref{fig:frames} shows qualitative comparisons between MVDepthNet, DeepMVS, COLMAP, and our MaskMVS approach. It should be noted that our method is the only method that can capture the small objects in the top left of the third row and the bottom left of the last row. Moreover, our method provide more accurate prediction for close areas (see the brightest parts of ground truths) in the first row and the fourth row.

\section{Discussion and Conclusions}
\label{sec:discussion}
We have proposed a novel CNN-based architecture for multi-view stereo depth estimation that is inspired by traditional plane-sweep algorithms without the need of constructing an explicit cost-volume metric. Instead, we designed a binary classification task for our MaskNet and used the mask-based multiplane representations to aggregate information from multiple views and exploit geometric relationships. Moreover, we discussed the effect of depth selection and proposed a novel way of sampling depth planes based on histogram matching. Our ablation study showed that uniformly sampling in the distribution domain can deal with both small depths such as indoor scenes and large depths such as in outdoor scenes, even with sparsely sampled planes. As the running time will drop when reducing the number of planes, our proposed sampling method can be beneficial for real-time systems that have restrictions on computation time and memory. Moreover, compared to traditional multi-view stereo methods, our approach can handle low-texture inputs and does not need iterative refinement; compared to other CNN-based methods that also employ a plane-sweep scheme, our method do not need to compute any distance metrics, which makes our method time-efficient. 

As ideal plane selection can lead to better prediction, one direction to improve our architecture might be adjusting depth planes according to inputs automatically. It should be noted that using predicted mask maps from our MaskNet, the depth distribution can be roughly estimated. Then it is possible to obtain uniform samples in distribution domain by just 1D linear interpolation. This might offer the possibility for varying sampled depth planes with different scenes in the future.
Codes are available at \url{https://github.com/AaltoVision/MaskMVS}.

{\bf Acknowledgements}~~We acknowledge computing resources by Aalto Science-IT and CSC, and funding from the Academy of Finland (308640 and 277685).

\clearpage
{\small
\bibliographystyle{splncs04}
\bibliography{bibliography}
}

\end{document}